\documentclass[11pt,a4paper]{article}
\usepackage[hyperref]{emnlp-ijcnlp-2019}
\usepackage{times}
\usepackage{latexsym}

\usepackage{url}
\usepackage{tikz}
\definecolor{gitred}{HTML}{FDB8C0}
\definecolor{gitgreen}{HTML}{006400}
\definecolor{chocolate}{HTML}{D2691E}
\definecolor{maroon}{HTML}{800000}
\definecolor{indigo}{HTML}{4B0082}
\definecolor{green}{HTML}{008000}
\usepackage{pifont}

\usepackage[normalem]{ulem}

\usepackage{enumitem}
\usepackage{booktabs}
\usepackage{multirow}
\usepackage{graphicx}
\usepackage{array}
\usepackage{subfigure}
\usepackage{amsmath}
\usepackage{url}
\usepackage{makecell}
\usepackage{amsmath}
\usepackage{amsfonts}
\usepackage{amsthm,amssymb,amsopn,bm}
\usepackage{color}
\usepackage{soul}
\usepackage{verbatim}

\usepackage{xcolor,colortbl}
\definecolor{green}{rgb}{0.1,0.1,0.1}
\usepackage{tabulary}
\usepackage{footnote}

\aclfinalcopy

\title{
    A Discrete Hard EM Approach for \\
    Weakly Supervised Question Answering
}

\author{Sewon Min$^1$, Danqi Chen$^{2,3}$, Hannaneh Hajishirzi$^{1,4}$, Luke Zettlemoyer$^{1,3}$ \\
$^1$University of Washington, Seattle, WA \\
$^2$Princeton University, Princeton, NJ \\
$^3$Facebook AI Research, Seattle, WA \\
$^4$Allen Institute for Artificial Intelligence, Seattle, WA \\
{\tt $\{$sewon,hannaneh,lsz$\}$@cs.washington.edu}
{\tt danqic@cs.princeton.edu}
}

\date{}

\begin{document}
\maketitle
\newcommand{\myabstract}{
    Many question answering (QA) tasks only provide weak supervision for how the answer should be computed. For example, \trivia\ answers are entities that can be mentioned multiple times in supporting documents, while \drop\ answers can be computed by deriving many different equations from numbers in the reference text.
    In this paper, we show it is possible to convert such tasks into discrete latent variable learning problems with a precomputed, task-specific set of possible {\em solutions} (e.g. different mentions or equations) that contains one correct option.
    We then develop a hard EM learning scheme that computes gradients relative to the most likely solution at each update. 
    Despite its simplicity, we show that this approach significantly outperforms previous methods on six QA tasks, including absolute gains of 2--10\%, and achieves the \sota\ on five of them.
    Using hard updates instead of maximizing marginal likelihood is key to these results as it encourages the model to find the one correct answer, which we show through detailed qualitative analysis.\footnote{Our code is publicly available at \url{https://github.com/shmsw25/qa-hard-em}.}
}

\newcommand{\dataurl}{\url{https://bit.ly/2HK1Fqn}}
\newcommand{\eg}{e.g.}
\newcommand{\discrete}{Reading comprehension with discrete reasoning}
\newcommand{\sota}{state-of-the-art}
\newcommand{\dev}{development}
\newcommand{\trivia}{\textsc{TriviaQA}}
\newcommand{\triviaopen}{\textsc{TriviaQA-open}}
\newcommand{\nqopen}{\textsc{NaturalQuestions-open}}
\newcommand{\narrative}{\textsc{NarrativeQA}}
\newcommand{\nq}{\textsc{Natural Questions}}
\newcommand{\drop}{\textsc{DROP}}
\newcommand{\wikisql}{\textsc{WikiSQL}}

\newcommand{\triviasize}{61,888 & 7,993 & 7,701}
\newcommand{\triviausize}{ 78,785 & 8,837 & 11,313}
\newcommand{\narrativesize}{32,747 & 3,461 & 10,557}
\newcommand{\nqsize}{79,168 & 8,757 & 3,610}
\newcommand{\dropsize}{ 46,973 & 5,850 & - }
\newcommand{\wikisqlsize}{ 56,355 & 8,421 & 15,878 }

\newcommand{\noname}{\textsc{$x \xrightarrow{} Z$}}
\newcommand{\ours}{Ours}
\newcommand{\triviafirstd}{48.6}
\newcommand{\triviammld}{47.0}
\newcommand{\triviaoursd}{50.7}
\newcommand{\nqfirstd}{23.6}
\newcommand{\nqmmld}{26.6}
\newcommand{\nqoursd}{28.8 }
\newcommand{\triviafirstt}{48.1}
\newcommand{\triviammlt}{47.4}
\newcommand{\triviaourst}{50.9}
\newcommand{\nqfirstt}{23.6}
\newcommand{\nqmmlt}{25.8}
\newcommand{\nqourst}{28.1}

\newcommand{\bert}{\textsc{BERT}}
\newcommand{\qanet}{\textsc{QANet}}

\newcommand{\hred}[2]{{\cellcolor[rgb]{1,#1,#1} #2}}
\newcommand{\hdred}[2]{\cellcolor[rgb]{#1,0,0} {\protect\color{white}{#2}}}
\newcommand{\tred}[1]{{\color[rgb]{1,0,0} {#1}}}
\newcommand{\tblue}[1]{{\color[rgb]{0,0,1} {#1}}}

\providecommand{\tspan}[1]{
    {\protect\color{purple}{#1}}
}
\providecommand{\tnumber}[1]{
    {\protect\color{purple}{\textbf{\texttt{#1}}}}
}
\begin{abstract}
    \myabstract
\end{abstract}

\section{Introduction}\label{sec:intro}\begin{figure}[!thb]
\centering
\resizebox{\columnwidth}{!}{\includegraphics[width=\textwidth]{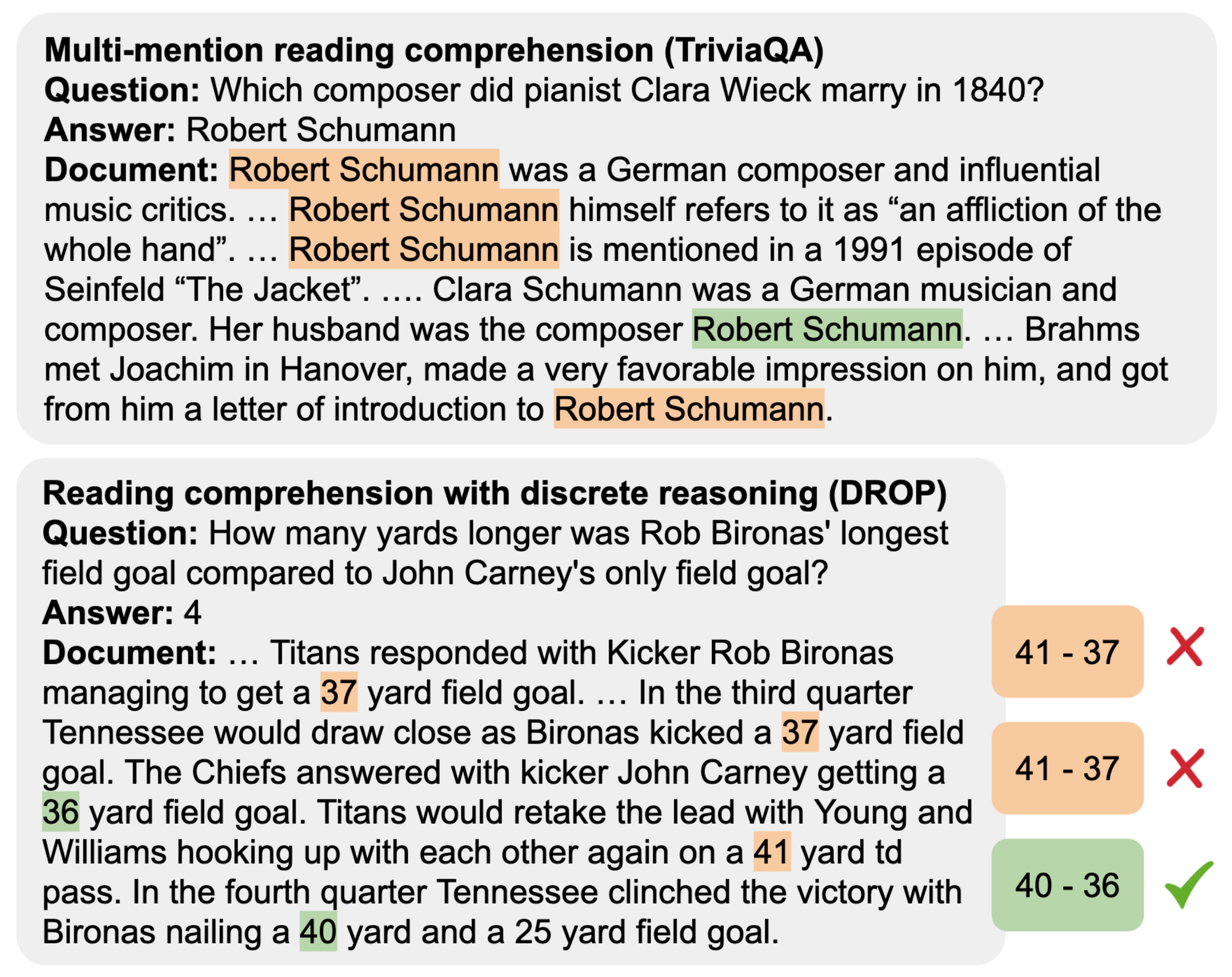}}
\caption{Examples from two different question answering tasks. \textbf{(Top) Multi-mention reading comprehension.} The answer text is mentioned five times in the given document, however, only the fourth span actually answers the question. \textbf{(Bottom) \discrete.} There are many potential equations which execute the answer (`4'), but only one of them is the correct equation (`40-36') and the others are false positives.
}
\label{fig:example}\end{figure}

\begin{table*}[tb]
    \centering \small
    \begin{tabulary}{\textwidth}{l|ccc|cc} 
     \toprule
        \multirow{2}{*}{Task \& Dataset} & \multicolumn{3}{c|}{\# Examples} & \multicolumn{2}{c}{$|Z|$} \\
        & Train & Dev & Test & Avg & Median \\
     \midrule
        \multicolumn{4}{l}{\textbf{1. Multi-mention reading comprehension}}
        \\
     \midrule
        \trivia{}~\citep{triviaqa} & \triviasize & 2.7 & 2 \\
        \narrative{}~\citep{narrativeqa} & \narrativesize & 4.3 & 5 \\
        \triviaopen~\citep{triviaqa} & \triviausize & 6.7 & 4 \\
        \nqopen~\citep{naturalquestions} & \nqsize & 1.8 & 1 \\
     \midrule
        \multicolumn{4}{l}{\textbf{2. \discrete}} \\
     \midrule
        \drop{}$_\text{num}$~\citep{drop}& \dropsize & 8.2 & 3 \\
     \midrule
        \multicolumn{4}{l}{\textbf{3. Semantic Parsing}} \\
     \midrule
        \wikisql{}~\citep{wikisql} & \wikisqlsize & 346.1 & 5 \\
     \bottomrule
\end{tabulary}
\caption{
    Six QA datasets in three different categories used in this paper (detailed in Section 5) along with the size of each dataset.
    An average and median of the size of precomputed solution sets (denoted by $Z$) are also reported.
    Details on how to obtain $Z$ are given in Section~\ref{sec:setup}.
} 
\label{tab:dataset}
\vspace{-8pt}
\end{table*}

A natural setting in many question answering (QA) tasks is to provide weak supervision to determine how the question should be answered given the evidence text. For example, as seen in Figure~\ref{fig:example},  \trivia\ answers are entities that can be mentioned multiple times in supporting documents, while \drop\ answers can be computed by deriving many different equations from numbers in the reference text. Such weak supervision is attractive because it is relatively easy to gather, allowing for large datasets, but complicates learning because there are  many different spurious ways to derive the correct answer. 
It is natural to model such ambiguities with a latent variable during learning, but most prior work on reading comprehension has rather focused on the model architecture and used heuristics to map the weak signal to full supervision (\eg\ by selecting the first answer span in \trivia\ ~\citep{triviaqa,tay2018densely,multiqa}). Some models are trained with maximum marginal likelihood (MML)~\citep{kadlec2016text,neural-cascades,clark2018multi,lee2019latent}, but it is unclear if it gives a meaningful improvement over the heuristics.

In this paper, we show it is possible to formulate a wide range of weakly supervised QA tasks as discrete latent-variable learning problems.
First, we define a {\em solution} to be a particular derivation of a model to predict the answer (\eg\ a span in the document or an equation to compute the answer).
We demonstrate that for many recently introduced tasks, which we group into three categories as given in Table~\ref{tab:dataset}, it is relatively easy to precompute a discrete, task-specific set of possible solutions that contains the correct solution along with a modest number of spurious options. 
The learning challenge is then to determine which solution in the set is the correct one, while estimating a complete QA model. 

We model the set of possible solutions as a discrete latent variable, and develop a learning strategy that uses hard-EM-style parameter updates. This algorithm repeatedly (i) predicts the most likely solution according to the current model from the precomputed set, and (ii) updates the model parameters to further encourage its own prediction.
Intuitively, these hard updates more strongly enforce our prior beliefs that there is a single correct solution. 
This method can be applied to any problem that fits our weak supervision assumptions and can be used with any model architecture. 

We experiment on six different datasets (Table~\ref{tab:dataset}) using strong task-specific model architectures~\citep{bert,drop,hwang2019comprehensive}. 
Our learning approach significantly outperforms previous methods which use heuristic supervision and MML updates, including absolute gains of 2--10\%, and achives the \sota\ on five datasets.
It outperforms recent \sota{} reward-based semantic parsing algorithms~\citep{mapo, mapox} by 13\% absolute percentage on \wikisql, strongly suggesting that having a small precomputed set of possible solutions is a key ingredient.
Finally, we present a detailed analysis showing that, in practice, the introduction of hard updates encourages models to assign much higher probability to the correct solution.

\section{Related Work}\label{sec:related}\paragraph{Reading Comprehension.} Large-scale reading comprehension (RC) tasks that provide full supervision for answer spans~\citep{squad} have seen significant progress recently~\citep{bidaf,dcn+,fast-and-accurate,bert}. 
More recently, the community has moved towards more challenging tasks such as distantly supervised RC~\citep{triviaqa}, RC with free-form human generated answers~\citep{narrativeqa} and RC requiring discrete or multi-hop reasoning~\citep{drop,hotpot}. 
These tasks introduce new learning challenges since the gold solution that is required to answer the question (\eg\ a span or an equation) is not given.

Nevertheless, not much work has been done for this particular learning challenge. Most work on RC focuses on the model architecture and simply chooses the first span or a random span from the document~\citep{triviaqa,tay2018densely,multiqa}, rather than modeling this uncertainty as a latent choice.
Others maximize the sum of the likelihood of multiple spans~\citep{kadlec2016text,neural-cascades,clark2018multi,lee2019latent}, but it is unclear if it gives a meaningful improvement.
In this paper, we highlight the learning challenge and show that our learning method, independent of the model architecture, can give a significant gain. Specifically, we assume that one of mentions are related to the question and others are false positives because (i) this happens for most cases, as the first example in Table~\ref{tab:formulation}, and (ii) even in the case where multiple mentions contribute to the answer, there is often a single span which fits the question the best.

\paragraph{Semantic Parsing.} Latent-variable learning has been extensively studied in the literature of semantic parsing~\citep{Zettlemoyer2005LearningTM,clarke2010driving,liang2013learning,berant2013semantic,artzi2013weakly}.
For example, a question and an answer pair ($x$, $y$) is given but the logical form that is used to compute the answer is not.
Two common learning paradigms are maximum marginal likelihood (MML) and reward-based methods. In MML, the objective maximizes $\sum_{z \in \hat{Z}}  \mathbb{P}(z|x)$, where $\hat{Z}$ is an approximation of a set of logical forms executing $y$~\citep{liang2013learning,berant2013semantic,krishnamurthy2017neural}.
In reward-based methods, a reward function is defined as a prior, and the model parameters are updated with respect to it~\citep{sequentialwikitable,imlfirst,mapo}.
Since it is computationally expensive to obtain a precomputed set in semantic parsing, these methods typically recompute the set of logical forms with respect to the beam at every parameter update.
In contrast, our learning method targets tasks that a set of solutions can be precomputed, which include many recent QA tasks such as reading comprehension, open-domain QA and a recent SQL-based semantic parsing task~\cite{wikisql}.

\section{Method}\label{sec:method}In this section, we first formally define our general setup, which we will instantiate for specific tasks in Section~\ref{sec:setup} and then we describe our learning approach.

\subsection{Setup}
Let $x$ be the input of a QA system (\eg\ a question and a document) and $y$ be the answer text (\eg\ `Robert Schumann' or `4').
We define a {\em solution} as a particular derivation that a model is supposed to produce for the answer prediction (\eg\ a span in the document or an equation to compute the  answer, see Table~\ref{tab:formulation}).
Let $f$ denote a task-specific, deterministic function which maps a solution to the textual form of the answer (\eg\ by simply returning the string associated with a particular selected mention or solving an equation to get the final number, see Table~\ref{tab:formulation}). Our goal is to learn a model (with parameters $\theta{}$) which takes an input $x$ and outputs a solution $z$ such that $f(z) = y$.

In a fully supervised scenario, a true solution $\bar{z}$ is given, and $\theta$ is estimated based on a collection of ($x, \bar{z}$) pairs. In this work, we focus on a weakly supervised setting in which $\bar{z}$ is not given and we define $Z_\text{tot}$ as a finite set of all the possible solutions. In the case that the search space is very large or infinite, we usually can approximate $Z_\text{tot}$ with a high coverage in practice. Then, we obtain $Z = \{ z \in Z_\text{tot}: f(z) = y \}$ by enumerating all $z \in Z_\text{tot}$.
This results a set of all the possible solutions that lead to the correct answer. We assume it contains one solution that we want to learn to produce, and potentially many other spurious ones. In practice, $Z$ is defined in a task-specific manner, as we will see in Section~\ref{sec:setup}.

At inference time, the model produces a solution $\tilde{z} \in Z_\text{tot}$ from an input $x$ with respect to $\theta$ and predicts the final answer as $f(\tilde{z})$.
Note that we cannot compute $Z$ at inference time because the groundtruth $y$ is not given.\footnote{
    This is a main difference from multi-instance learning (MIL)~\citep{multiinstancelearning}, since a bag of input instances is given at inference time in MIL.
}

\subsection{Learning Method}
In a fully-supervised setting where $\bar{z}$ is given, we can learn $\theta$ by optimizing the negative log likelihood of $\bar{z}$ given the input $x$ with respect to $\theta{}$.
\begin{align*}
    J_{\text{Sup}}(\theta{}|x, \bar{z}) = - \mathrm{log} ~\mathbb{P}(\bar{z} | x; \theta{})
\end{align*}
In our weak supervision scenario, the model has access to $x$ and $Z = \{ z_1, z_2, \dots, z_n \}$, and the selection of the best solution in $Z$ can be modeled as a latent variable. We can compute the maximum marginal likelihood (MML) estimate, which marginalizes the likelihood of each $z_i \in Z$ given the input $x$ with respect to $\theta{}$. Formally, 
\begin{eqnarray*}
    \mathbb{P}(y|x; \theta{}) &=& \sum_{z_i \in Z_\text{tot}} \mathbb{P}(y|z_i) \mathbb{P}(z_i|x; \theta{}) \\
    &=& \sum_{z_i \in Z} \mathbb{P}(z_i|x; \theta{})
\end{eqnarray*} is used to compute the objective as follows:
\begin{equation*}
    J_{\text{MML}}(\theta{}|x, Z) = - \mathrm{log} \sum_{z_i \in Z}  \mathbb{P}(z_i | x; \theta{})
\end{equation*}
However, there are two major problems in the MML objective in our settings. 
First, MML can be maximized by assigning high probabilities to any subset of $z_i \in Z$; whereas in our problems, instances in $Z$ other than one correct $z$ are spurious solutions which the model should ideally assign very low probability.
Second, in MML we optimize the sum over probabilities of $Z$ during training but typically predict the maximum probability solution during inference, creating a discrepancy between training and testing.

\begin{table*}[tb]
    \centering \scriptsize
    \begin{tabulary}{\textwidth}{l} 
    \toprule
        \textbf{1. Multi-Mention Reading Comprehension} (\trivia{}, \narrative{}, \triviaopen\ \& \nqopen) \\
    \midrule
        \textbf{Question:} Which composer did pianist Clara Wieck marry in 1840? \\
        \textbf{Document:} \tspan{Robert Schumann} was a German composer and influential music critic. He is widely regarded as one of the greatest composers of the Romantic \\
        era. (...) \tspan{Robert Schumann} himself refers to it as ``an affliction of the whole hand". (...) \tspan{Robert Schumann} is mentioned in a 1991 episode of Seinfeld ``The \\
        Jacket". (...) Clara Schumann was a German musician and composer, considered one of the most distinguished pianists of the Romantic era. Her husband was \\ the composer \tspan{Robert Schumann}. $<$Childhood$>$ (...) At the age of eight, the young Clara Wieck performed at the Leipzig home of Dr. Ernst Carus. There \\ she met another gifted young pianist who had been invited to the musical evening, named \tspan{Robert Schumann}, who was nine years older. Schumann admired \\ Clara' s playing so much that he asked permission from his mother to discontinue his law studies. (...) In the spring of 1853, the then unknown 20-year-old \\Brahms met Joachim in Hanover, made a very favorable impression on him, and got from him a letter of introduction to \tspan{Robert Schumann}. \\
        \textbf{Answer ($\boldsymbol{y}$):} Robert Schumann \\
        \textbf{$\boldsymbol{f}$:} Text match \\
        \textbf{$\boldsymbol{Z_\text{tot}}$}: All spans in the document \\
        \textbf{$\boldsymbol{Z}$:} Spans which match `Robert schumann' (red text) \\
    \midrule
        \textbf{2. Reading Comprehension with Discrete Reasoning} (\drop{}$_\text{num}$) \\
    \midrule
        \textbf{Question:} How many yards longer was Rob Bironas' longest field goal compared to John Carney's only field goal? \\
        \textbf{Document:} (...) The Chiefs tied the game with QB Brodie Croyle completing a \tnumber{10} yard td pass to WR Samie Parker. Afterwards the Titans responded with \\ Kicker Rob Bironas managing to get a \tnumber{37} yard field goal. Kansas city would take the lead prior to halftime with croyle completing a \tnumber{9} yard td pass to FB \\ Kris Wilson. In the third  quarter Tennessee would draw close as Bironas kicked a \tnumber{37} yard field goal. The Chiefs answered with kicker John Carney getting \\ a \tnumber{36} yard field goal. Afterwards the Titans would retake the lead with Young and Williams hooking up with each other again on a \tnumber{41} yard td pass. \\ (...) Tennessee clinched the victory with Bironas nailing a \tnumber{40} yard and a \tnumber{25} yard field goal. With the win the Titans kept their playoff hopes alive at \tnumber{8}\tnumber{6}. \\
        \textbf{Answer ($\boldsymbol{y}$):} 4 \\
        \textbf{$\boldsymbol{f}$:} Equation executor \\
        \textbf{$\boldsymbol{Z_\text{tot}}$}: Equations with two numeric values and one arithmetic operation \\ \textbf{$\boldsymbol{Z}$:} \{ 41-37, 40-36, 10-6,  ... \} \\
    \midrule
        \textbf{3. SQL Query Generation} (\wikisql{}) \\
    \midrule
        \textbf{Question:} What player played guard for Toronto in 1996-1997? \\ \textbf{Table Header:} \texttt{player, year, position, ...}\\
        \textbf{Answer ($\boldsymbol{y}$):} John Long \\
        \textbf{$\boldsymbol{f}$:} SQL executor \\
        \textbf{$\boldsymbol{Z_\text{tot}}$:} Non-nested SQL queries with up to 3 conditions\\
        \textbf{$\boldsymbol{Z}$:} \texttt{Select {player} where position=guard and year in toronto=1996-97} \\
        \texttt{Select max(player) where position=guard and year in toronto=1996-97} \\ \texttt{Select min(player) where position=guard} \\
        \texttt{Select min(player) where year in toronto=1996-97} \\
        \texttt{Select min(player) where position=guard and year in toronto=1996-97} \\
     \bottomrule
\end{tabulary}
\caption{
    Examples of the input, answer text ($y$), $f$, $Z_\text{tot}$ and $Z$. First, in multi-mention reading comprehension, the answer text `Robert Schumann' is mentioned six times but only the fourth span is related to the question. Second, in reading comprehension with discrete reasoning, many equations yield to the answer 4, but only `40-37' answers the question. Lastly, in SQL query generation, five SQL queries lead to the answer but only the first one is the correct query.
    See Section~\ref{sec:setup} for more details.
} 
\label{tab:formulation}
\vspace{-8pt}
\end{table*}

We introduce a learning strategy with a hard-EM approach. First, the model computes the likelihood of each $z_i$ given the input $x$ with respect to $\theta{}$, $\mathbb{P}(z_i | x; \theta{})$, and picks one of $Z$ with the largest likelihood:
\begin{equation*}
    \tilde{z} = \mathrm{argmax}_{z_i \in Z} \mathbb{P} (z_i | x; \theta{})
\end{equation*}
Then, the model optimizes on a standard negative log likelihood objective, assuming $\tilde{z}$ is a true solution. The objective can be re-written as follows:
\begin{eqnarray*}
    J_{\text{Hard}}(\theta{}|x, Z)
    &=& - \mathrm{log}~\mathbb{P}(\tilde{z} | x; \theta{}) \\
    &=& - \mathrm{log}~\mathrm{max}_{z_i \in Z}  \mathbb{P}(z_i | x; \theta{}) \\
    &=& - \mathrm{max}_{z_i \in Z} \mathrm{log} ~\mathbb{P}(z_i | x; \theta{}) \\
    &=& \mathrm{min}_{z_i \in Z} J_{\text{Sup}}(\theta{}|x, z_i)
\end{eqnarray*}

\noindent
This objective can be seen as a variant of MML, where the $\mathrm{sum}$ is replaced with a $\mathrm{max}$.

\section{Task Setup}\label{sec:setup}
We apply our approach to three different types of QA tasks: multi-mention reading comprehension, RC with discrete reasoning and a semantic parsing task. In this section, we describe each task in detail: how we define a solution $z$ and pre-compute a set $Z$ based on input $x$ and answer $y$. The statistics of $|Z|$ and examples on each task are shown in Table~\ref{tab:dataset} and Table~\ref{tab:formulation} respectively.

\subsection{Multi-Mention Reading Comprehension}\label{subsec:dsr}
Multi-mention reading comprehension naturally occurs in several QA tasks such as (i) distantly-supervised reading comprehension where a question and answer are collected first before the evidence document is gathered (\eg\ \trivia), (ii) abstractive reading comprehension which requires a free-form text to answer the question (\eg\ \narrative), and (iii) open-domain QA where only question-answer pairs are provided.

Given a question $Q = [q_1, \dots, q_l]$ and a document $D = [d_1, \dots, d_L]$, where $q_i$ and $d_j$ denote the tokens in the question and document, the output $y$ is an answer text, which is usually mentioned multiple times in the document.

Previous work has dealt with this setting by detecting spans in the document through text matching~\citep{triviaqa, clark2018multi}.
Following previous approaches, we define a solution $z$ as a span in the document. We obtain a set of possible solutions $Z = \{ z_1, \dots, z_n \}$ by finding exact match or similar mentions of $y$, where $z_i=(s_i, e_i)$ is a span of text with start and end token indices $s_i$ and $e_i$. Specifically,
\begin{eqnarray*}
    g_\text{max} &=& \max\limits_{1 \leq s_i \leq e_i \leq L }  g([d_{s_i}, \dots, d_{e_i}],  y) \\
    Z &=& \{z_i=(s_i, e_i)~\text{s.t. } g(s_i, e_i) = g_\text{max}\},
\end{eqnarray*}
where $g$ is a string matching function. If the answer is guaranteed to be a span in the document $D$, $g$ is a binary function which returns $1$ if two strings are the same, and $0$ otherwise. If the answer is free-form text, we choose $g$ as the ROUGE-L metric~\citep{lin2004rouge}.

This complicates the learning because the  given document contains many spans matching to the text, while most of them are not related to answering the question. As an example shown in Table~\ref{tab:formulation}, only the fourth span out of six is relevant to the question.

\subsection{Reading Comprehension with Discrete Reasoning}\label{subsec:drop-setup}
Some reading comprehension tasks require reasoning in several discrete steps by finding clues from the document and aggregating them. One such example is mathematical reasoning, where the model must pick numbers from a document and compute an answer through arithmetic operations~\cite{drop}. 

In this task, the input is also a question $Q$ and a document $D$, and the output $y$ is given as a numeric value. We define a solution $z$ to be an executable arithmetic equation. Since there is an infinite set of potential equations, we approximate $Z_\text{tot}$ as a finite set of arithmetic equations with two numeric values and one operation, following~\citet{drop}.\footnote{This approximation covers 93\% of the examples in the \dev\ set.} Specifically, 
\begin{align*}
    {Z}_\text{tot} = \big\{ &z_i = (o_1, n_1, o_2, n_2) ~\text{s.t.} \\ 
    &o_1, o_2 \in \{ +, -, \% \}, \\
    &n_1, n_2 \in N_D \cup N_Q \cup S \big\},
\end{align*}
where $N_D$ and $N_Q$ are all numeric values appearing in $D$ and $Q$, respectively, and $S$ are a set of predefined special numbers. Then
\begin{equation*}
    Z = \{ z_i \in {Z}_\text{tot}~\text{s.t. } f(z_i) = y
    \}
\end{equation*}
where $f$ is an execution function of equations.

Figure~\ref{fig:example} shows an example $Z$ given a question and a document. We can see that one equation is correct, while the others are false positives which coincidentally lead to the correct answer.

\subsection{SQL Query Generation}
To evaluate if our training strategy generalizes to other weak supervision problems, we also study a semantic parsing task where a question and an answer are given but the logical form to execute the answer is not. In particular, we consider a task of answering questions about a given table by generating SQL queries.

The input is a question $Q = [q_1, \dots, q_l]$ and a table header $H = [h_1, \dots, h_{n_L}]$, where $q_i$ is a token, $h_i$ is a multi-token title of each column, and $n_L$ is the number of headers. The supervision is given as the SQL query result $y$, which is always a text string.

We define a solution to be an SQL query.
Since the set of potential queries is infinite, we approximate $Z_\text{tot}$ as a set of non-nested SQL queries with at most three conditions.\footnote{This approximation covers 99\% of the examples in the \dev\ set.} 
Specifically, given $A$ as a set of aggregating operators $\{ \text{sum}, \text{mean}, \text{max}, \text{min}, \text{count} \}$ and $C$ as a set of possible conditions $\{ (h, o, t)~\text{s.t. } h \in [1, n_L], o \in \{ =,<,> \}, t \in \text{spans in } Q \}$, we define ${Z}_\text{tot}$:
\begin{eqnarray*}
    {Z}_\text{tot} = \{ z_i &=& (z_i^\text{sel}, z_i^\text{agg}, \{ z_{i,j}^\text{cond}\}_{j=1}^3 \} ~\text{s.t.} \\
    z_i^\text{sel} &\in& [1, n_L] \\
    z_i^\text{agg} &\in& \{\text{none}\} \cup A \\
    z_{i,j}^\text{cond} &\in& \{\text{none}\} \cup C~\text{for } j \in [1, 3]\}, 
\end{eqnarray*}
then,
\begin{eqnarray*}
    Z = \{ z_i  \in {Z}_\text{tot}~\text{s.t. } f(z_i) = y 
    \}, 
\end{eqnarray*}
where $f$ is an SQL executor. The third example in Table~\ref{tab:formulation} shows ${Z}$ may contain many spurious SQL querie, \eg\ the third query in ${Z}$ coincidentally executes the answer because `John Long' is ranked first among all the guards in alphabetical order.

\section{Experiments}\label{sec:exp}We experiment on a range of question answering tasks with varied model architectures to demonstrate the effectiveness of our approach. Built on top of strong base models, our learning method is able to achieve \sota{} on \narrative, \triviaopen, \nqopen, \drop$_\text{num}$ and \wikisql.

\begin{table*}[tb]
    \centering \small
    \begin{tabulary}{\textwidth}{l|cc|cc|cc|cc|c|c} 
     \toprule
        & \multicolumn{2}{c|}{\multirow{2}{*}{\trivia{}}} 
        & \multicolumn{2}{c|}{\multirow{2}{*}{\narrative{}}} 
        & \multicolumn{2}{c|}{\trivia{}}
        & \multicolumn{2}{c|}{\textsc{NaturalQ}}
        & {\drop{}$_\mathrm{num}$}
        & {\drop{}$_\mathrm{num}$}
        \\
        & & & &
        & \multicolumn{2}{c|}{\textsc{-open}} & \multicolumn{2}{c|}{\textsc{-open}} & {w/ BERT} & {w/ QANet} \\
        & \multicolumn{2}{c|}{{(F1)}}
        & \multicolumn{2}{c|}{{(ROUGE-L)}}
        & \multicolumn{2}{c|}{(EM)}
        & \multicolumn{2}{c|}{(EM)}
        & (EM) & (EM) \\
        & Dev & Test & Dev & Test & Dev & Test & Dev & Test & Dev & Dev \\
     \midrule
        First Only  & 64.4 & 64.9 & 55.3 & 57.4 & \triviafirstd & \triviafirstt &\nqfirstd & \nqfirstt & 42.9 & 36.1 \\
        MML         & 64.8 & 65.5 & 55.8 & 56.1 & \triviammld & \triviammlt & \nqmmld & \nqmmlt & 39.7 & 43.8 \\
        \ours       & 66.9 & 67.1 & \textbf{58.1} & \textbf{58.8}& \textbf{\triviaoursd} & \textbf{\triviaourst} & \textbf{\nqoursd} & \textbf{\nqourst} & \textbf{52.8} & \textbf{45.0} \\
      \midrule
        {SOTA} & - & \textbf{71.4} & - & 54.7 & 47.2 & 47.1 & 24.8 &  26.5 & \multicolumn{2}{c}{43.8} \\
     \bottomrule
\end{tabulary}
\caption{
    \textbf{Results on multi-mention reading comprehension \& discrete reasoning tasks.} We report performance on five datasets with different base models.
    Note that we are not able to obtain the test result on the subset \drop{}$_\mathrm{num}$. Previous \sota{} are from \citet{wang2018multi}, \citet{nishida2019multi}, \citet{lee2019latent}, \citet{lee2019latent} and \citet{drop}, respectively.
    Our training method consistently outperforms the First-Only and MML by a large margin in all the scenarios.
} 
\label{tab:rc-result}
\vspace{-8pt}
\end{table*}

\begin{table}[!tb]
    \centering \footnotesize
    \begin{tabulary}{\textwidth}{lcc} 
        \toprule
            Model & \multicolumn{2}{c}{Accuracy} \\
            & Dev & Test \\
        \midrule
            \multicolumn{2}{l}{{\em{Weakly-supervised setting}}} \\
        \midrule
            REINFORCE~\citep{reinforce} & $< 10$ & \\
            Iterative ML~\citep{imlfirst} & 70.1 &\\
            Hard EM~\citep{mapo} & 70.2 &\\
            Beam-based MML~\citep{mapo} & 70.7 &\\
            MAPO~\citep{mapo} & 71.8 & 72.4 \\
            MAPOX~\citep{mapox} & 74.5 & 74.2 \\
            MAPOX+MeRL~\citep{mapox} & 74.9 & 74.8 \\
            MML & 70.6 & 70.5 \\
            \ours & \textbf{84.4} & \textbf{83.9} \\
        \midrule
            \multicolumn{2}{l}{{\em{Fully-supervised setting}}} \\
        \midrule
            SQLNet~\citep{sqlnet} & 69.8 & 68.0 \\
            TypeSQL~\citep{typesql} & 74.5 & 73.5\\
            Coarse2Fine~\citep{dong2018coarse} & 79.0 & 78.5 \\
            SQLova~\citep{hwang2019comprehensive} & 87.2 & 86.2\\
            X-SQL~\citep{xsql} & \textbf{89.5} & \textbf{88.7}\\
        \bottomrule
    \end{tabulary}
    \caption{
        \textbf{Results on \wikisql.} We compare accuracy with weakly-supervised or fully-supervised settings.
        Our method outperforms previous weakly-supervised methods and most of published fully-supervised methods. 
    }
    \label{tab:wikisql-result}
\vspace{-8pt}
\end{table}

\subsection{Multi-mention Reading Comprehension}\label{subsec:multi-mention-result}

We experiment on two reading comprehension datasets and two open-domain QA datasets. For reading comprehension, we evaluate on \trivia{} (Wikipedia)~\citep{triviaqa} and \narrative{} (summary)~\citep{narrativeqa}.

For open-domain QA, we follow the settings in \citet{lee2019latent} and use the QA pairs from \trivia-unfiltered~\citep{triviaqa} and \nq~\citep{naturalquestions} with short answers and discard the given evidence documents. We refer to these two datasets as \triviaopen\ and \nqopen.\footnote{
Following \citet{lee2019latent}, we treat the dev set as the test set and split the train set into 90/10 for training and development.
Datasets and their split can be downloaded from \dataurl.}

We experiment with three learning methods as follows.
\vspace{-2pt}
\begin{itemize}\itemsep -.7pt
    \item First Only: $J(\theta{}) = - \mathrm{log} \mathbb{P}(z_1|x;\theta{})$, where $z_1$ appears first in the given document among all $z_i \in Z$.
    \item MML: $J(\theta{}) = - \mathrm{log} \Sigma_{i=1}^{n}  \mathbb{P}(z_i|x;\theta{})$.
    \item Ours: $J(\theta{}) = - \mathrm{log} \mathrm{max}_{1 \leq i \leq n}  \mathbb{P}(z_i|x;\theta{})$.
\end{itemize}

$\mathbb{P}(z_i|Q, D)$ can be obtained by any model which outputs the start and end positions of the input document. In this work, we use a modified version of \bert{}~\citep{bert} for multi-paragraph reading comprehension~\citep{min2019compositional}.

\paragraph{Training details.} We use uncased version of \bert{}$_\text{base}$.
For all datasets, we split  documents into a set of segments up to 300 tokens because \bert{} limits the size of the input.
We use batch size of $20$ for two reading comprehension tasks and $192$ for two open-domain QA tasks.
Following \citet{clark2018multi}, we filter a subset of segments in \trivia\ through TF-IDF similarity between a segment and a question to maintain a reasonable length. 
For open-domain QA tasks, we retrieve 50 Wikipedia articles through TF-IDF~\citep{squad-open} and further run BM25~\citep{robertson2009probabilistic} to retrieve 20 (for train) or 80 (for development and test) paragraphs. We try 10, 20, 40 and 80 paragraphs on the development set to choose the number of paragraphs to use on the test set.

To avoid local optima, we perform annealing: at training step $t$, the model optimizes on MML objective with a probability of $\mathrm{min} (t/\tau{}, 1)$ and otherwise use our objective, where $\tau{}$ is a hyperparameter. We observe that the performance is improved by annealing while not being overly sensitive to the hyperparameter $\tau{}$. We include full hyperparameters and detailed ablations in Appendix~\ref{app:annealing}.

\paragraph{Results.}
Table~\ref{tab:rc-result} compares the results of baselines, our method and the \sota{} on four datasets.\footnote{For \narrative, we compare with models trained on \narrative\ only. For open-domain QA, we only compare with models using pipeline approach.} First of all, we observe that First-Only is a strong baseline across all the datasets. We hypothesize that this is due to the bias in the dataset that answers are likely to appear earlier in the paragraph.
Second, while MML achieves comparable result to the First-Only baseline, our learning method outperforms others by $2+$ F1/ROUGE-L/EM consistently on all datasets.
Lastly, our method achieves the new \sota{} on \narrative{}, \triviaopen\ and \nqopen, and is comparable to the \sota{} on \trivia{}, despite our aggressive truncation of documents.

\subsection{Reading Comprehension with Discrete Reasoning}\label{subsec:descrete-reasoning-result}
We experiment on a subset of \drop~\citep{drop} with numeric answers (67\% of the entire dataset) focusing on mathematical reasoning. We refer to this subset as  \drop{}$_\text{num}$. The current \sota{} model is an augmented version of \qanet{}~\cite{fast-and-accurate} which selects two numeric values from the document or the question and performs addition or subtraction to get the answer. The equation to derive the answer is not given, and \citet{drop} adopted the MML objective.

$\mathbb{P}(z_i|Q, D)$ can take as any model which generates equations based on the question and document. Inspired by \citet{drop}, we take a sequence tagging approach on top of two competitive models: (i) augmented \qanet{}, the same model as \citet{drop} but only supporting addition, subtraction and counting, and (ii) augmented \bert{}, which supports addition, subtraction and percentiles.\footnote{As we use a different set of operations for the two models, they are not directly comparable. Details of the model architecture are shown in Appendix~\ref{app:model-details}.} 

\paragraph{Training details.}
We truncate the document to be up to 400 words. 
We use the batch size of $14$ and $10$ for \qanet\ and \bert, respectively.

\paragraph{Results.}
Table~\ref{tab:rc-result} shows the results on \drop{}$_\text{num}$. Our training strategy outperforms the First-Only baseline and MML by a large margin, consistently across two base models. In particular, with \bert{}, we achieve an absolute gain of 10\%. 

\begin{figure*}[!t]
\centering
\subfigure[DROP$_\text{num}$]{
\includegraphics[trim=5 5 5 5,clip,width=.42\textwidth]{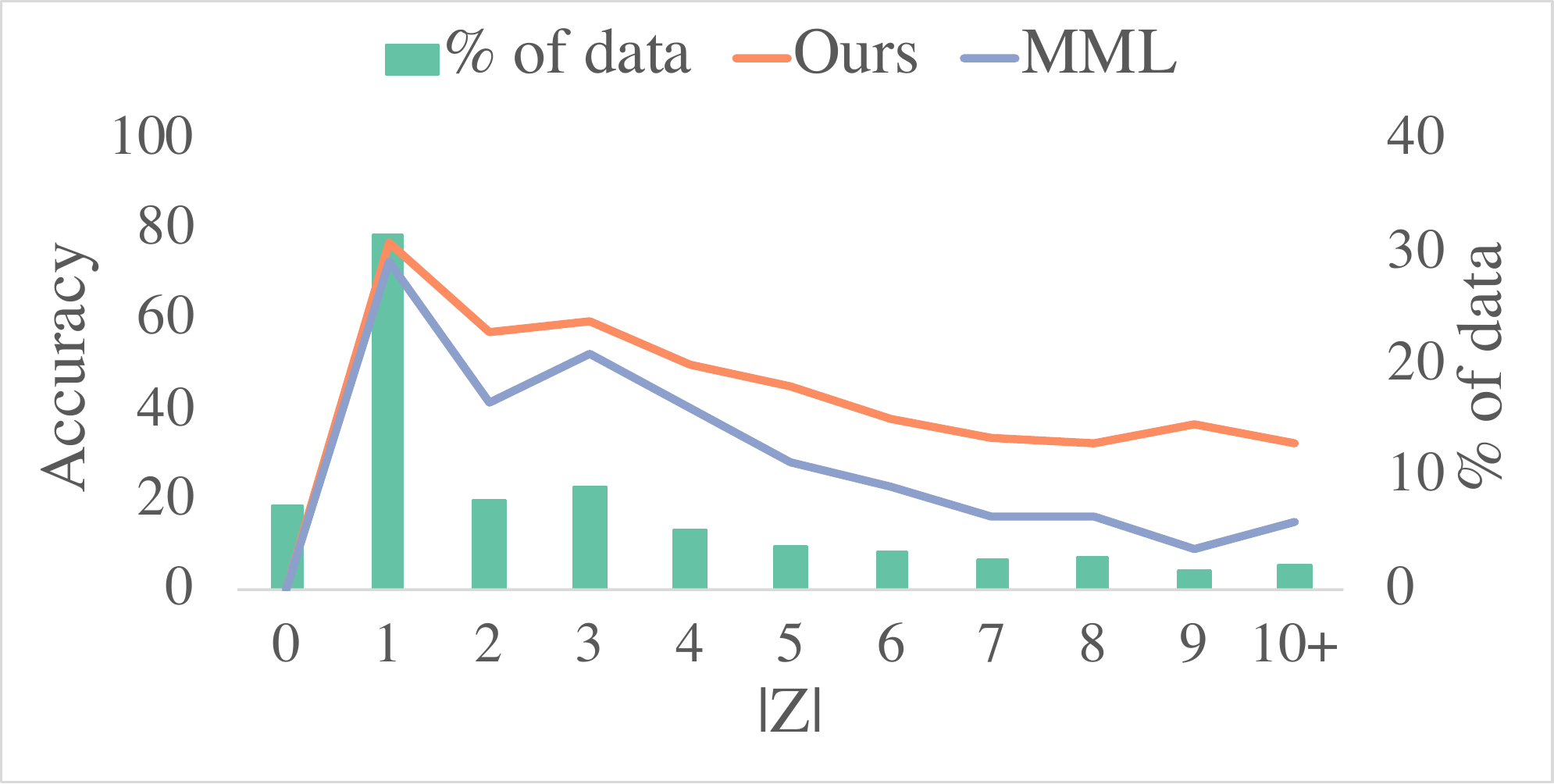}} 
\hspace{1em}
\subfigure[WikiSQL]{
\includegraphics[trim=5 5 5 5,clip,width=.525\textwidth]{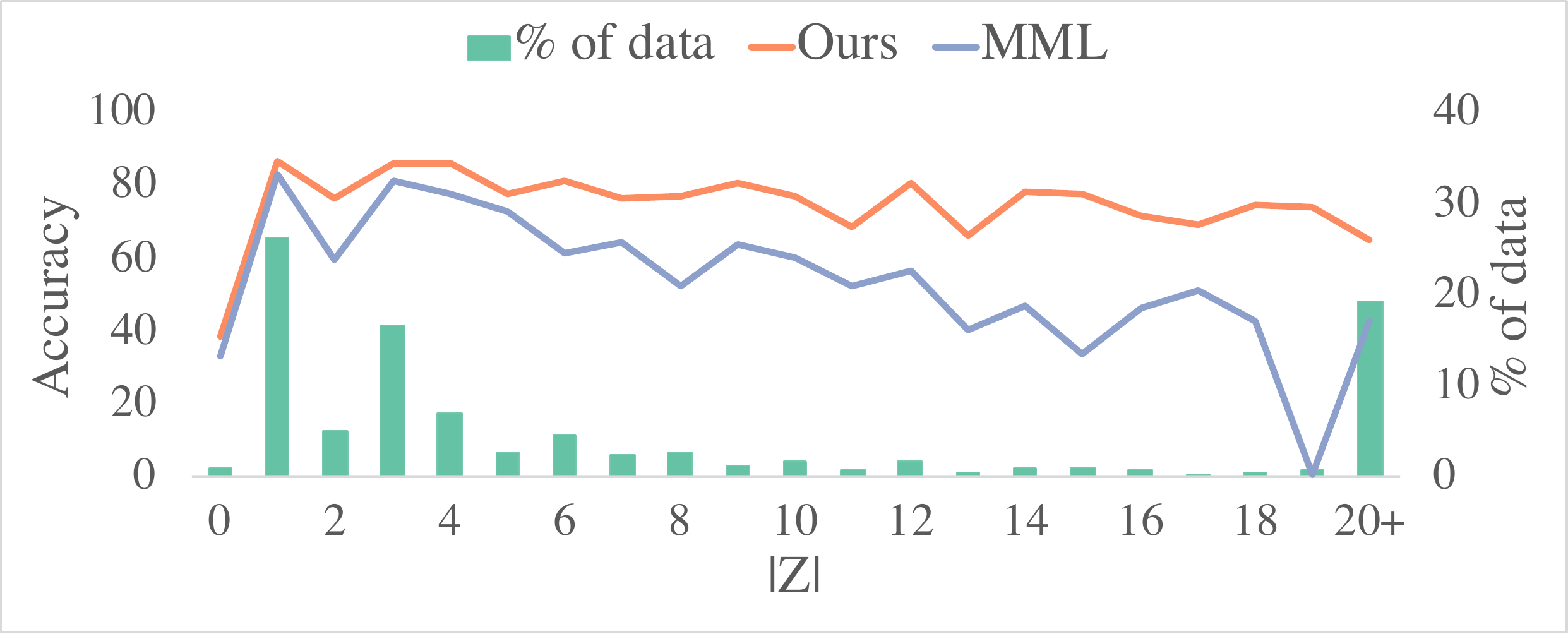}}
\caption{
    \textbf{Varying the size of solution set ($\boldsymbol{|Z|}$) at test time.} We compare the model trained on MML objective (blue) and our training strategy (orange). Our approach consistently outperforms MML on \drop$_\text{num}$ and \wikisql, especially when $|Z|$ is large.
}
\label{fig:breakdown}
\end{figure*}
\begin{figure*}[!t]
\centering \footnotesize
\begin{tabulary}{\textwidth}{l|ccc} 
    \toprule
        Group & Avg $|Z|$ & Median $|Z|$ & \# train \\
    \midrule
        3 & 3.0 & 3 & 10k  \\
        10 & 10.2 & 9 & 10k \\
        30 & 30.0 & 22 & 10k \\
        100 & 100.6 & 42 & 10k\\
        300 & 300.0 &  66 & 10k\\
    \bottomrule
\end{tabulary}
\hspace{2em}
\raisebox{-0.4\totalheight}{\subfigure{
\includegraphics[trim=5 5 5 5,clip,width=.5\textwidth]{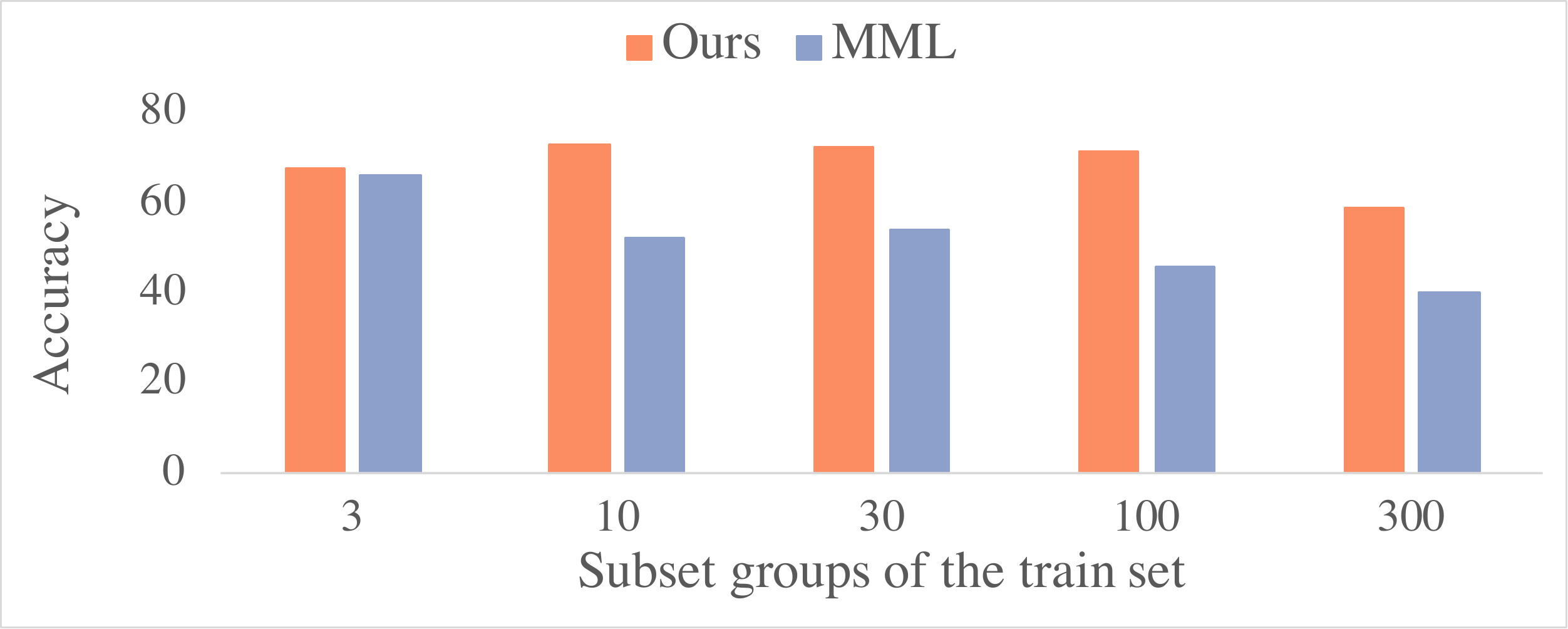}}}
\caption{
    \textbf{Varying the size of solution set ($\boldsymbol{|Z|}$) at training.} {(Left)} Subsets of the train set on \wikisql{} varying in the size of solution set ($|Z|$). All subsets contain 10k training examples (total in the original train set is 55k). All subsets are evaluated on the same, original \dev\ set for a fair comparison.
    {(Right)} Performance across subsets of the training set with varying $|Z|$. Our method achieves substantial gains over MML.
}
\label{fig:ablation}
\end{figure*}

\subsection{SQL Query Generation}\label{subsec:semantic-parsing-result}

Finally, we experiment on the weakly-supervised setting of \wikisql~\citep{wikisql}, in which only the question \& answer pair is used and the SQL query $(z)$ is treated as a latent variable. 

$\mathbb{P}(z_i|Q, H)$ can be computed by any query generation or semantic parsing models. We choose SQLova~\citep{hwang2019comprehensive}, a competitive model on \wikisql~(designed for fully supervised setting), as our base model. We modify the model to incorporate either the MML objective or our hard-EM learning approach for the weakly-supervised setting.

We compare with both traditional and recently-developed reward-based algorithms for weak supervision, including beam-based MML (MML which keeps a beam during training), conventional hard EM\footnote{This method differs from ours in that it does not have a precomputed set, and uses a beam of candidate predictions to execute for each update.}, REINFORCE~\citep{reinforce}, iterative ML~\citep{imlfirst,imlsecond} and a family of MAPO (Memory-augmented policy optimization)~\citep{mapo,mapox}. For a fair comparison, we only consider single models without execution-guided decoding.

\paragraph{Training details.}
We adopt the same set of hyperparameters as in \citet{hwang2019comprehensive}, except that we change the batch size to 10 and truncate the input to be up to 180 words. 

\paragraph{Results.}
Table~\ref{tab:wikisql-result} shows that our training method significantly outperforms all the weakly-supervised learning algorithms, including 10\% gain over the previous state of the art.
These results indicate that precomputing a solution set and training a model through hard updates play a significant role to the performance.
Given that our method does not require SQL executions at training time (unlike MAPO), it provides a simpler, more effective and time-efficient strategy. Comparing to previous models with full supervision, our results are still on par and outperform most of the published results. 

\section{Analysis}\label{sec:analysis}In this section, we will conduct thorough analyses and ablation studies, to better understand how our model learns to find a solution from a precomputed set of possible solutions. We also provide more examples and analyses in Appendix~\ref{app:wikisql-examples}.

\begin{table*}[!tb]
\centering \footnotesize
\begin{tabular}{l|}
    \textbf{Q:} How many yards longer was Rob Bironas' longest field goal compared \\
    to John Carney's only field goal? (\textbf{Answer:} 4) \\
    \textbf{P:} ... The Titans responded with Kicker Rob Bironas managing to get a 37\\
    yard field goal. ...Tennessee would draw close as Bironas kicked a 37 yard \\
    field goal. The Chiefs answered with kicker John Carney getting a 36 yard \\
    field goal. The Titans would retake the lead with Young and Williams hook-\\
    ing up with each other again on a 41 yard td pass. ...Tennessee clinched the\\
    victory with Bironas nailing a 40 yard and a 25 yard field goal. \\
\end{tabular}
\setlength\tabcolsep{1.5pt}
\begin{tabular}{c|c|c|c|c|c}
    $t$ & Pred & \multicolumn{4}{c}{$Z$ (ordered by $\mathbb{P}(z|x; \theta_{t})$)} \\
    \hline 
    {$1$k} &
    \hred{0.45}{10-9} & \hred{0.71}{10-6} & \hred{0.88}{41-37} & \hred{0.88}{40-36} & \hred{0.89}{41-37$^\ddagger$}\\
    {$2$k} &
    \hdred{0.98}{37-36} & \hred{0.37}{40-36} & \hred{0.75}{41-37} & \hred{0.78}{41-37$^\ddagger$} & \hred{1}{10-6}\\
    {$4$k} &
    \hred{0.31}{40-36} & \hred{0.31}{40-36} & \hred{0.83}{41-37$^\ddagger$} & \hred{0.85}{41-37} & \hred{1}{10-6}\\
    {$8$k} &
    \hred{0.42}{40-36} & \hred{0.42}{40-36} & \hred{0.74}{41-37$^\ddagger$} & \hred{0.84}{41-37} & \hred{1}{10-6}\\
    {$16$k} &
    \hred{0.31}{37-36} & \hred{0.39}{40-36} & \hred{0.78}{41-37} & \hred{0.96}{41-37$^\ddagger$} & \hred{1}{10-6}\\
    {$32$k} &
    \hdred{0.63}{40-36} & \hdred{0.63}{40-36} & \hred{0.92}{41-37} & \hred{0.98}{41-37$^\ddagger$} & \hred{1}{10-6}\\
\end{tabular}
\caption{
    An example from \drop$_\text{num}$ (same as Figure~\ref{fig:example} and Table~\ref{tab:formulation}), with its answer text `4' and a subset of the solution set ($Z$), containing two of `41-38' (which `41' come from different mentions; one denoted by $\ddagger$ for distinction), `40-36' and `10-4'.
    For each training step $t$, the top 1 prediction and $Z$ ordered by $P(z|x; \theta_{t})$, a probability of $z \in Z$ with respect to the model at $t$ through training procedure are shown.
    Note that at inference time $Z$ is not given, so top 1 prediction is not necessarily an element of $Z$.
} 
\label{tab:drop-vis}
\vspace{-8pt}
\end{table*}

\paragraph{\textbf{Varying the size of solution set at inference time.}} Figure~\ref{fig:breakdown} shows a breakdown of the model accuracy with respect to the size of a solution set ($|Z|$) at test time.
We observe that the model with our training method outperforms the model with MML objective consistently across different values of $|Z|$.
The gap between MML and our method is marginal when $|Z|=0$ or $1$, and gradually increases as $|Z|$ grows.

\paragraph{\textbf{Varying the size of solution set at training.}} To see how our learning method works with respect to the size of a solution set ($|Z|$) of the training data, particularly with large $|Z|$, we take 5 subsets of the training set on \wikisql{} with $|Z|=3,10,30,100, 300$. We train a model with those subsets and evaluate it on the original \dev\ set, both with our training method and MML objective. Figure~\ref{fig:ablation} shows statistics of each subset and results. We observe that (i) our learning method outperforms MML consistently over different values of $|Z|$, and (ii) the gain is particularly large when $|Z|>3$.

\paragraph{Model predictions over training.} We analyze the top 1 prediction and the likelihood of $z \in Z$ assigned by the model on \drop$_\text{num}$ with different number of training iterations (steps from 1k to 32k). 
Table~\ref{tab:drop-vis} shows one example on \drop$_\text{num}$ with the answer text `4', along with the model's top 1 prediction and a subset of $Z$.
We observe that the model first begins by assigning a small, uniform probability distribution to $Z$, but gradually learns to favor the true solution. The model sometimes gives the wrong prediction---for example, at $t=16$k, and changes its prediction from the true solution to the wrong solution, `37-36'---but again changes its prediction to be a true solution afterward. In addition, its intermediate wrong solution, `37-36' indicates the model was confused with distinguishing the longest field goal of Rob Bironas (40 vs. 37), which is an understandable mistake.

We also compare the predictions from the model with our method to those from the model with MML, which is shown in Appendix~\ref{app:wikisql-examples}.

\paragraph{Quality of the predicted solution.}

We analyze if the model outputs the correct solution, since the solution executing the correct answer could be spurious.
First, on \narrative\ and \drop$_\text{num}$, we manually analyze 50 samples from the \dev\ set and find that 98\% and 92\% of correct cases produce the correct solution respectively.
Next, on \wikisql, we compare the predictions from the model to the annotated SQL queries on the \dev\ set. This is possible because gold SQL queries are available in the dataset for the full supervision. Out of 8,421 examples, 7,110 predictions execute the correct answers. Among those, 88.5\% of the predictions are exactly same as the annotated queries. Others are the cases where (i) both queries are correct, (ii) the model prediction is correct but the annotated query is incorrect, and (iii) the annotated query is correct and the model prediction is spurious. We show a full analysis in Appendix~\ref{app:wikisql-examples}.

\paragraph{Robustness to the noise in $\boldsymbol{|Z|}$.}

Sometimes noise arises during the construction of $|Z|$, such as $|Z|$ constructed based on ROUGE-L for \narrative. 
To explore the effect of noise in $Z$, we experiment with more noisy solution set by picking all the spans with scores that is equal to or larger than the 5th highest. The new construction method increases $|Z|$ from 4.3 to 7.1 on \narrative.
The result by MML objective drops significantly (56.07$\rightarrow{}$51.14) while the result by ours drops marginally (58.77$\rightarrow{}$57.97), suggesting that MML suffers more with a noisier $Z$ while ours is more robust.

\section{Conclusion}\label{sec:concl}In this paper, we demonstrated that, for many QA tasks which only provide the answer text as supervision, it is possible to precompute a discrete set of possible solutions that contains one correct option.
Then, we introduced a discrete latent variable learning algorithm which iterates a procedure of predicting the most likely solution in the precomputed set and further increasing the likelihood of that solution.
We showed that this approach significantly outperforms previous approaches on six QA tasks including reading comprehension, open-domain QA, discrete reasoning task and semantic parsing, achieving absolute gains of 2--10\% and setting the new \sota\ on five well-studied datasets.

\subsection*{Acknowledgements}
This research was supported by ONR (N00014-18-1-2826, N00014-17-S-B001), DARPA N66001-19-2-403, NSF (IIS-1616112, IIS-1252835, IIS-1562364), ARO (W911NF-16-1-0121), an Allen Distinguished Investigator Award, Samsung GRO and gifts from Allen Institute for AI, Google and Amazon. 

The authors would like to thank the anonymous reviewers, Eunsol Choi, Christopher Clark, Victor Zhong and UW NLP members for their valuable feedback.

\bibliography{journal-abbrv,bib}
\bibliographystyle{acl_natbib}

\clearpage
\appendix
\section{Model details}\label{app:model-details}We describe the detailed model architecture used as a base model. In other words, we describe how we obtain $\mathbb{P}(z|x; \theta)$.

The following paragraphs describe (i) \bert{} extractive model used for multi-mention RC (Section~\ref{subsec:multi-mention-result}) and (ii) \bert{} sequence tagging model for discrete reasoning task (Section~\ref{subsec:descrete-reasoning-result}), respectively. For \qanet{} for discrete reasoning and the model for SQL generation (Section~\ref{subsec:semantic-parsing-result}), we use the open-sourced code of the original implementation\footnote{\url{https://github.com/allenai/allennlp/blob/master/allennlp/models/reading_comprehension/naqanet.py} and \url{https://github.com/naver/sqlova}} of~~\citet{drop} and~\citet{hwang2019comprehensive} and do not make any modification except the objective function, so we refer to original papers.

All implementations are done in Pytorch~\citep{pytorch}. 
For \bert{}, we modify the open-sourced implementation in PyTorch\footnote{\url{https://github.com/huggingface/pytorch-pretrained-BERT}} and use the uncased version of \bert{}$_\text{base}$.

\paragraph{Extractive QA model for multi-mention RC}
The model architecture is closed to that of~\citet{min2019compositional} and~\citet{alberti2019bert}, where the model operates independently on each paragraph, and selects the best matching paragraph and its associated answer span.

The input is a question $Q$ and a set of paragraphs $\{P_1, \dots, P_N\}$, and the desired output is a span from one of paragraphs. Since our goal is to compute a probability of a specific span, $z$, let's say $z$ is $s$-th through $e$-th word in $k$-th paragraph.

The model receives a question $Q$ and a single paragraph $P_i$ in parallel. Then, $S_i = Q:{\tt [SEP]}:P$, a list of $m+n_i+1$ words, where : indicates a concatenation, ${\tt [SEP]}$ is a special token, $m$ is the length of $Q$, and $n_i$ is the length of $P_i$. This $S_i$ is fed into~\bert:
\begin{equation*}
    \bar{S}_i = \bert (S) \in \mathbb{R}^{h \times (m+n_i+1)}
\end{equation*}
where $h$ is the hidden dimension of~\bert. Then, 
\begin{eqnarray*}
    p_{i,\textrm{start}} &=& \mathrm{Softmax}  \big( \bar{S}_i^T W_1  \big)  \in \mathbb{R}^{m+n_i+1} \\
    p_{i,\textrm{end}} &=& \mathrm{Softmax} \big( \bar{S}_i^T W_2 \big) \in \mathbb{R}^{m+n_i+1} 
\end{eqnarray*}
where $W_1, W_2 \in \mathbb{R}^{h}$ are learnable vectors.

Finally, the probability of $z$, $s$-th through $e$-th word in $i$-th paragraph, is obtained by:
\begin{equation*}
    \mathbb{P}(z|Q, P_i) = p_{i,\textrm{start}}^s \times p_{i,\textrm{end}}^e
\end{equation*}
where $p^d$ denotes $d$-th element of the vector $p$.

Separately, a paragraph selector is trained through $p_{i,\textrm{exit}} = \mathrm{Softmax} \big( W_3 \mathrm{maxpool}(\bar{S}_i) \big) \in \mathbb{R}^2$ where $W_3 \in \mathbb{R}^{h \times 2}$ is learnable vector. At inference time, $k = \mathrm{argmax}_i p_{i,\textrm{exit}}^1$ is computed and $\mathbb{P}(z|Q, P_k)$ is only considered to output a span.


\paragraph{Sequence Tagging model for discrete reasoning}
The basic idea of the model is closed to that of~\citet{drop}.
The input is a question $Q$ and a paragraph $P$. Our goal is to compute a probability of an equation, $z = (o_1, n_1, o_2, n_2)$, where $ o_1, o_2 \in \{ +, -, *0.01 \}$ and $n_1, n_2 \in N_P \cup N_Q \cup S$, $N_P$ and $N_Q$ are all numeric values appearing in $P$ and $Q$, and $S$ are a set of predefined special numbers.\footnote{$S=\{1,2,3,4,5,7,10,12,100,1000\}$}.

First, \bert{} encodings of the question and the paragraph is obtained via \begin{equation*}
    \bar{S} = \bert (Q:{\tt [SEP]}:P) \in \mathbb{R}^{h \times (m+n+1)}
\end{equation*}
where : indicates a concatenation, ${\tt [SEP]}$ is a special token, $m$ is the length of $Q$, $n$ is the length of $P$, and $h$ is the hidden dimension of~\bert. Then, 
\begin{eqnarray*}
    p_\text{input} &=& \mathrm{Softmax} \big( \bar{S}^T W_1  \big) \in \mathbb{R}^{(m+n+1) \times 4} \\
    p_\text{special} &=& \mathrm{Softmax} \big( \mathrm{maxpool}(\bar{S}) W_2 \big) \in \mathbb{R}^{|S| \times 4}
\end{eqnarray*}
where $W_1 \in \mathbb{R}^{h \times 4}$ and $W_2 \in \mathbb{R}^{h \times |S| \times 4}$ are learnable matrices. Then,
\begin{eqnarray*}
    \mathbb{P}(z|x) = \Pi_{i=1}^{m+n+1} p_\text{input}^{g(i)} \Pi_{j=1}^{|S|} p_\text{special}^{h(j)}
\end{eqnarray*}
where
\begin{eqnarray*}
    g(i) &=&  \begin{cases}
                \alpha(o_1)  & \text{if } [Q:[\texttt{SEP}]:P]^i=n_1 \\
                \alpha(o_2)  & \text{if } [Q:[\texttt{SEP}]:P]^i=n_2 \\
                0       & \text{o.w.}
            \end{cases} \\
    h(j) &=&  \begin{cases}
                \alpha(o_1)  & \text{if } S^j=n_1 \\
                \alpha(o_2)  & \text{if } S^j=n_2 \\
                0       & \text{o.w.}
            \end{cases} \\
    \alpha(o) &=&  \begin{cases}
                1  & \text{if } o=\text{`+'} \\
                2  & \text{if } o=\text{`-'} \\
                3  & \text{if } o=\text{`*0.01'}
            \end{cases}
\end{eqnarray*}
Here, subscript $i$ of the vector or the sequence indicate $i$-th dimension of the vector or $i$-th element of the sequence, respectively.

\begin{table*}[!t]
\centering \footnotesize
\begin{tabulary}{\textwidth}{l|c} 
    \toprule
        $\tau$ & \trivia~F1 \\
    \midrule
        None & 55.83 \\
        20k & 58.05 \\
        30k & 57.99 \\
        40k & 56.66 \\
    \bottomrule
\end{tabulary}
\hspace{2em}
\begin{tabulary}{\textwidth}{l|c} 
    \toprule
        $\tau$ & \drop$_\text{num}$~EM  \\
    \midrule
        None & 52.31 \\
        5k & 51.98 \\
        10k & 52.82 \\
        20k & 51.74 \\
    \bottomrule
\end{tabulary}
\hspace{2em}
\begin{tabulary}{\textwidth}{l|c|c} 
    \toprule
        Dataset & batch size & $\tau$ \\
    \midrule
        \trivia & 20 & 15K  \\
        \narrative & 20 & 20K \\
        \triviaopen & 192 & 4K \\
        \nqopen & 192 & 8K \\
        \drop$_\text{num}$ with \bert & 14 & 10K \\
        \drop$_\text{num}$ with \qanet & 14 & None \\
        \wikisql & 10 & None \\
    \bottomrule
\end{tabulary}
\caption{
    \textbf{(Left)} Ablations with varying values of $\tau$ on \trivia.
    \textbf{(Middle)} Ablations with varying values of $\tau$ on \drop$_\text{num}$ with \bert.
    \textbf{(Right)} Final $\tau$ chosen for the main results on each dataset. Note that for \drop$_\text{num}$ with \qanet\ and \wikisql, we just report the number without annealing.
}
\label{tab:annealing}
\end{table*}

\section{Annealing}\label{app:annealing}
To prevent the model to be optimized on early decision of the model, we perform annealing: at training step $t$, the model optimizes on MML objective with a probability of $\mathrm{min} (t/\tau{}, 1)$ and otherwise use our objective, where $\tau{}$ is a hyperparameter. We observe that the performance is improved by annealing while not being sensitive to the hyperparameter $\tau{}$. Ablations and chosen $\tau$ for each dataset are shown in Table~\ref{tab:annealing}. Note that for \drop$_\text{num}$ with \qanet\ and \wikisql, we do not ablate with varying $\tau$ and just report the number without annealing.

\section{Examples}\label{app:wikisql-examples}\begin{table*}[!tb]
\centering \footnotesize	
\begin{tabular}{l}
    Question \& Document \\
    \hline
    \textbf{Q:} What is the state capital of Alabama? (Groundtruth: Montgomery) \\
    \textbf{D:} Alabama is nicknamed the Yellowhammer State, after the state bird. Alabama is also known as the ``Heart of Dixie" and \\
    the ``Cotton State". The state tree is the longleaf pine, and the state flower is the camellia. Alabama's capital is \underline{\tred{Montgomery}}. \\ (...) From 1826 to 1846, \tblue{Tuscaloosa} served as Alabama's capital. On January 30, 1846, the Alabama legislature announced \\
    it had voted to move the capital city from Tuscaloosa to \underline{Montgomery}. The first legislative session in the new capital met in \\
    December 1847. A new capitol building was erected under the direction of Stephen Decatur Button of Philadelphia. The first \\
    structure burned down in 1849, but was rebuilt on the same site in 1851. This second capitol building in \underline{Montgomery} remains \\
    to the present day. \\
\end{tabular}
\caption{
    An example from \trivia{} with multiple spans of the answer text (\underline{underlined}). The model trained with self-training technique outputs the correct answer (\tred{red}) and the model trained on MML objective does not (\tblue{blue}). 
} 
\label{tab:trivia-vis}
\vspace{-8pt}
\end{table*}

\begin{table*}[!tb]
\centering \footnotesize
\begin{tabular}{l}
    Question \& Passage \\
    \hline
    \textbf{Q:} How many sports are not olympic sports but are featured in the asian games ? (A: 10) \\
    \textbf{P:} The first 30 sports were announced by the singapore national olympic council on 10 december 2013 on the sidelines of \\
    the 27th sea games in myanmar. It announced then that there was room for as many as eight more sports. On 29 april 2014 \\
    the final six sports namely boxing equestrian floorball petanque rowing and volleyball were added to the programme. \\
    Floorball will feature in the event for the first time after being a demonstration sport in the 2013 edition. In its selection of \\
    events the organising committee indicated their desire to set a model for subsequent games in trimming the number of `trad- \\
    itional' sports to refocus on the seag' s initial intent to increase the level of sporting excellence in key sports. Hence despite \\
    room for up to eight traditional sports only two floorball and netball were included in the programme. Amongst the other 34 \\
    sports 24 are olympic sports and all remaining sports are featured in the asian games. 
\end{tabular}
\setlength\tabcolsep{1.5pt} 
\begin{tabular}{c|c|c|c|c|c}
    \hline 
    \multicolumn{6}{c}{\textbf{\em Ours}} \\
    \hline
    $t$ & Pred & \multicolumn{4}{c}{$Z$ (ordered by $\mathbb{P}(z|x; \theta_{t})$)} \\
    \hline 
    {$1$k} &
    \hred{0.54}{10+two} & \hred{0.56}{eight+two} & \hred{0.62}{eight+two$^\ddagger$} & \hred{0.81}{34-24} & \hred{0.84}{10}\\
    {$2$k} &
    \hred{0.37}{30-24} &  \hred{0.37}{34-24} & \hred{0.80}{eight+two}  & \hred{0.89}{eight+two$^\ddagger$}  & \hred{0.9}{10}\\
    {$4$k} &
    \hred{0.12}{34+24} & \hred{0.12}{34-24} & \hred{0.81}{eight+two}  & \hred{0.83}{eight+two$^\ddagger$}  & \hred{0.89}{10}\\
    {$8$k} &
    \hred{0.18}{34+24} & \hred{0.18}{34-24} & \hred{0.71}{eight+two}  & \hred{0.92}{10} & \hred{0.99}{eight+two$^\ddagger$} \\
    {$16$k} &
    \hdred{0.76}{34-24} & \hdred{0.76}{34-24} & \hred{0.94}{eight+two}  & \hred{0.96}{eight+two$^\ddagger$} & \hred{1.0}{10} \\
    {$32$k} &
    \hdred{0.32}{34-24} & \hdred{0.32}{34-24} & \hred{0.94}{eight+two}  & \hred{0.95}{eight+two$^\ddagger$} & \hred{1.0}{10}\\
\end{tabular}
\begin{tabular}{c|c|c|c|c|c}
    \hline 
    \multicolumn{6}{c}{\textbf{\em MML}} \\
    \hline 
    $t$ & Pred & \multicolumn{4}{c}{$Z$ (ordered by $\mathbb{P}(z|x; \theta_{t})$)} \\
    \hline 
    {$1$K} &
    \hred{0.79}{10+two} & \hred{0.82}{eight+two} & \hred{0.82}{eight+two$^\ddagger$} & \hred{0.90}{10} & \hred{0.92}{34-24} \\
    {$2$k} &
    \hred{0.75}{eight+two} &  \hred{0.75}{eight+two} & \hred{0.77}{eight+two$^\ddagger$} & \hred{0.89}{34-24}  & \hred{0.9}{10}\\
    {$4$k} &
    \hred{0.74}{24+5}  &  \hred{0.75}{eight+two} & \hred{0.76}{eight+two$^\ddagger$} & \hred{0.89}{34-24}  & \hred{0.92}{10}\\
    {$8$k} &
    \hred{0.68}{24+5} & \hred{0.8}{eight+two} & \hred{0.79}{eight+two$^\ddagger$} & \hred{0.93}{34-24}  & \hred{0.94}{10}\\
    {$16$k} &
    \hred{0.72}{34-24}  & \hred{0.72}{34-24}  &  \hred{0.87}{eight+two} & \hred{0.88}{eight+two$^\ddagger$} & \hred{0.94}{10}\\
    {$32$k} &
    \hred{0.72}{24+5}  & \hred{0.81}{34-24}  &  \hred{0.87}{eight+two} & \hred{0.88}{eight+two$^\ddagger$} & \hred{0.93}{10}\\
\end{tabular}
\caption{
    An example from \drop{}$_\text{num}$, with its answer text `10' and a subset of $Z$, containing `10', two of `eight+two' (which `eight' come from different mentions; one denoted by `$\ddagger$' for distinction) and `34-24'.
    The below tables are predictions from the model with our training strategy (left) and MML (right).
    For each training step $t$, the top 1 prediction and $Z$ ordered by $P(z|x; \theta_{t})$, a probability of $z \in Z$ with respect to the model at $t$ are shown.
    Note that at inference time $Z$ cannot be obtained, so top 1 prediction is not necessarily  in $Z$.
} 
\label{tab:drop-vis-comparison}
\vspace{-8pt}
\end{table*}

\begin{table*}[!tb]
    \centering \small
    \begin{tabulary}{\textwidth}{cl} 
        \toprule
            \textbf{Q} & How many times was the \# of total votes 2582322? \\
            \textbf{H} & Election, \# of candidates nominated, \# of seats won, \# of total votes, \% of popular vote \\
            \textbf{A} & \texttt{Select count(\# of seats won) where \# of total votes = 2582322} \\
            \textbf{P} & \texttt{Select count(Election) where \# of total votes = 2582322} \\
        \midrule
            \textbf{Q} & What official or native languages are spoken in the country whose capital city is Canberra? \\
            \textbf{H} & Country (exonym), Capital (exonym), Country (endonym) Capital (endonym), Official or native language \\
            \textbf{A} & \texttt{Select Official or native languages where} \texttt{Capital (exonym) = Canberra} \\
            \textbf{P} & \texttt{Select Official or native languages where} \texttt{Capital (endonym) = Canberra} \\
        \midrule
            \textbf{Q} & What is the episode number that has production code 8abx15? \\
            \textbf{H} & No. in set, No. in series, Title, Directed by, Written by, Original air date, Production code \\
            \textbf{A} & \texttt{Select min(No.in series) where} \texttt{Production code = 8ABX15} \\
            \textbf{P} & \texttt{Select No.in series where} \texttt{Production code = 8abx15}  \\
        \midrule
            \textbf{Q} & what is the name of the battleship with the battle listed on May 13, 1915? \\
            \textbf{H} & Estimate, Name, Nat., Ship Type, Principal victims, Date \\
            \textbf{A} & \texttt{Select Name where Ship Type = battleship and Date = may 13, 1915} \\
            \textbf{P} & \texttt{Select Name where Date = may 13, 1915} \\
        \bottomrule
    \end{tabulary}
    \caption{
        Four examples from \wikisql{} where the prediction from the model is different from annotated SQL query, although the executed answers are the same. \textbf{Q}, \textbf{H}, \textbf{A} and \textbf{P} indicate the given question, the given table header, annotated SQL query and predicted SQL query. First two example shows the case where both queries are correct. Next example shows the case that the model prediction makes more sense than the annotated query. The last example shows the cases that the annotated query makes  more sense than the model prediction.
    }
    \label{tab:wikisql-examples}
\vspace{-8pt}
\end{table*}

To see if the prediction from the model is the correct solution to derive the answer, we analyze outputs from the model.

\paragraph{\trivia.} Table~\ref{tab:trivia-vis} shows one example from \trivia\  where the answer text (Montgomery) is mentioned in the paragraph multiple times. Predictions from the model with our training method and that with MML objective are shown in the red text and the blue text, respectively. The span predicted by the model with our method actually answers to the question, while other spans with the answer text is not related to the question.

\paragraph{\drop$\boldsymbol{_\text{num}}$.} Table~\ref{tab:drop-vis-comparison} shows predictions from the model with our method and that with MML objective over training procedure. We observe that the model with our method learns to assign a high probability to the best solution (`34-24'), while the model with MML objective fails to do so.
Another notable observation is that the model with our method assign sparse distribution of likelihood over $Z$, compared to the model with MML objective. We quantitatively define sparsity as \begin{align*}\frac{|\{z \in Z~\text{s.t. } \mathbb{P}(z|x; \theta)<\epsilon\}|}{|Z| }\end{align*}\citep{hurley2009comparing} and show that the model with our method gives higher sparsity than the model with MML (59 vs. 36 with $\epsilon=10^{-3}$, 54 vs. 17 with $\epsilon=10^{-4}$ on \drop).

\paragraph{\wikisql.} \wikisql\ provides the annotated SQL queries, makes it easy to compare the predictions from the model to the annotated queries. Out of 8421 examples from the development set, 7110 predictions execute the correct answers. Among those, 6296 predictions are exactly same as the annotated queries. For cases where the predictions execute the correct answers but are not exactly same as the groundtruth queries, we show four examples in Table~\ref{tab:wikisql-examples}. In the first example, both the annotated query and the prediction are correct, because the selected column does not matter for counting. Similarly in the second example, both queries are correct because \texttt{Capital (exonym)} and \texttt{Capital (endonym)} both indicate the capital city. In the third example, the prediction makes more sense than the annotated query because the question does not imply anything  about \texttt{min}. In the last example, the annotated query makes more sense than the prediction because the prediction misses \texttt{Ship Type = battleship}. We conjecture that the model might learn to ignore some information in the question if the table header implies the table is specific about that information, hence does not need to condition on that information.

\end{document}